%% file: main.tex
\pdfoutput=1

\documentclass[11pt]{article}

\usepackage[utf8]{inputenc} 
\usepackage[T1]{fontenc}    
\usepackage{hyperref}       
\usepackage{url}            
\usepackage{booktabs}       
\usepackage{amsfonts}       
\usepackage{nicefrac}       
\usepackage{microtype}      
\usepackage{amssymb}
\usepackage{amsbsy}
\usepackage{amsmath}
\usepackage{amsthm}
\usepackage{latexsym}
\usepackage{subcaption}
\usepackage{wrapfig}
\usepackage{array}
\usepackage{colortbl}
\usepackage{balance} 
\usepackage{graphicx}
\usepackage{ulem}
\usepackage[font=small]{caption}
\usepackage{algorithm}
\usepackage{algorithmicx, algpseudocode}
\usepackage{mathtools}
\usepackage{arydshln}
\usepackage{titlesec}
\usepackage{multirow}
\usepackage[dvipsnames]{xcolor} 
\usepackage{acl}
\usepackage{graphicx}
\usepackage{times}
\usepackage{latexsym}
\usepackage{booktabs}
\usepackage{multirow}
\usepackage{hyperref}
\usepackage{url}
\usepackage{amsmath}
\usepackage{amssymb}
\usepackage{mathtools}
\usepackage{amsthm}
\usepackage{thmtools, thm-restate}
\usepackage{bbm}
\usepackage{nccmath}
\usepackage{tabularx}
\newcommand{\method}{\texttt{InfoPO}\xspace}
\usepackage{amssymb}
\usepackage{amsmath}
\usepackage{xspace}
\usepackage{booktabs}
\usepackage{caption}
\usepackage{subcaption}
\usepackage{inconsolata}
\newtheorem{theorem}{Theorem}[section]

\usepackage{adjustbox}
\usepackage{xcolor}
\definecolor{urlcolor}{HTML}{3333A6}
\usepackage{listings}
\newcommand\blfootnote[1]{%
  \begingroup
  \renewcommand\thefootnote{}\footnote{#1}%
  \addtocounter{footnote}{-1}%
  \endgroup
}

\newcolumntype{M}[1]{>{\small\arraybackslash}m{#1}}

\usepackage{todonotes}

\usepackage[dvipsnames]{xcolor} 
\title{InfoPO: On Mutual Information Maximization  for \\ Large Language Model Alignment}

\author{Teng Xiao$^\dagger$, Zhen Ge, Sujay Sanghavi, Tian Wang, Julian Katz-Samuels, \\
\textbf{Marc Versage}, \textbf{Qingjun Cui}, \textbf{Trishul Chilimbi} \\
Amazon\\
\texttt{tengxiao@psu.edu, sanghavi@mail.utexas.edu}\\
\texttt{\{zge, wangtan, jkatzsam, sversage, qingjunc, trishulc\}@amazon.com} 
}

\begin{document}
\maketitle

\begin{abstract}
We study the post-training of large language models (LLMs) with human preference data. Recently, direct preference optimization and its variants have shown considerable promise in aligning language models, eliminating the need for reward models and online sampling. Despite these benefits, these methods rely on explicit assumptions about the Bradley-Terry (BT) model, which makes them prone to overfitting and results in suboptimal performance, particularly on reasoning-heavy tasks. To address these challenges, we propose a principled preference fine-tuning algorithm called InfoPO, which effectively and efficiently aligns large language models using preference data. InfoPO eliminates the reliance on the BT model and prevents the likelihood of the chosen response from decreasing. Extensive experiments confirm that InfoPO consistently outperforms established baselines on widely used open benchmarks, particularly in reasoning tasks.
\end{abstract}

\input{sections/introduction}

\input{sections/related}

\input{sections/method}

\input{sections/experiments}



\balance
\bibliography{custom}
\bibliographystyle{acl_natbib}

\input{sections/appendix-proof}
\end{document}

%% file: sections/introduction.tex
\section{Introduction}
\blfootnote{$^\dagger$ Work done during internship at Amazon}
Large language Model alignment with human preferences is critical to ensure that the responses of pre-trained LLMs to prompts are consistent with human preferences~\citep{bai2022training, ouyang2022training, stiennon2020learning}. Recently, Reinforcement Learning from Human Feedback (RLHF) \citep{ouyang2022training, christiano2017deep} has been proposed for fine-tuning language models based on human preferences. RLHF involves initially fitting a reward signal derived from human preference data with the application of reinforcement learning  algorithms, such as Proximal Policy Optimization \citep{schulman2017proximal}, to optimize language model policy to maximize rewards.

\begin{figure*}[t!]
\centering 
\includegraphics[width=0.999\textwidth]{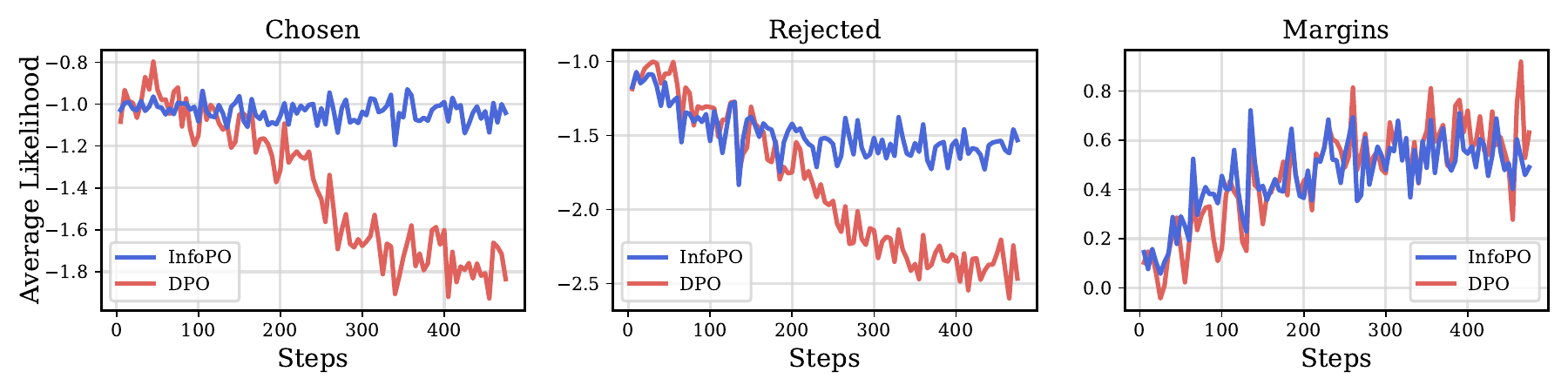}
\vskip -1em
\caption{The training dynamics of average likelihood  of \method and \texttt{DPO} on the Mistral-7B. We observe that \method exhibits the less decline in average chosen likelihoods, while still achieving the significant increase in  margins of rejected and chosen likelihood, compared to \texttt{DPO}. Results on Llama3-8B are given in Section~\ref{exp:anly}.}
\vskip -1em
\label{fig:rewards-mistral} 
\end{figure*}

RLHF demonstrates impressive capabilities across diverse tasks. Yet, the reinforcement learning approach presents significant challenges, such as computational inefficiency and training instability \citep{engstrom2020implementation, rafailov2024direct}. To address these issues, methods such as direct preference optimization and its variants have been proposed, including DPO \citep{rafailov2024direct}, R-DPO \citep{park2024disentangling}, and SimPO~\citep{meng2024simpo}. These preference optimization approaches \citep{tajwar2024preference}, replace RLHF with supervised learning on preference data, eliminating the need for explicit reward modeling. Specifically, they use the \textit{likelihood} of a policy to define an \textit{implicit reward} fitted to the preference data, achieving promising alignment performance.

While these methods employ different losses, they are all based on BT assumption and share a similar motivation with DPO and SimPO: maximizing the \textit{relative} value differences between the implicit rewards of the chosen and rejected responses. Despite its simplicity and initial promise, this BT assumption may not always hold true and  generally
decreases reasoning task performance, as discussed in~\cite{pal2024smaug,meng2024simpo,xiaocal}. Specifically, a notable counter-intuitive observation is that during the  training process of methods with BT assumption, the likelihood of both the chosen (i.e., preferred) and rejected (i.e., less preferred) responses decreases due to the large gradient associated with the rejected response. This leads to an undesirable outcome where the learned policy progressively focuses on unlearning the rejected responses (see section~\ref{sec:method} for details) and decrease the likelihood of chosen responses as shown in Figure~\ref{fig:rewards-mistral}, resulting in suboptimal performance on  reasoning benchmarks as shown in many recent works~\citep{xu2024dpo,meng2024simpo,pang2024iterative,chen2024noise}. Recently, several efforts~\cite{xu2024contrastive,pal2024smaug} have
been made to address this issue. They  propose using negative log likelihood  regularization on chosen responses to stabilize the training process. Although these methods successfully prevent the model from collapsing on mathematical datasets, they perform poorly on instruction-following  and chat benchmarks~\cite{meng2024simpo} and introduce additional hyper-parameters which require manual tuning. 

The importance of keeping the likelihood of the chosen response in practical applications of large language models, such as reasoning and mathematical problem-solving \citep{pal2024smaug, yuan2024advancing}, highlights a significant limitation in the applicability of contrastive preference learning. This raises the following question: \textit{Can we design an effective preference optimization algorithm that avoids the Bradley-Terry (BT) assumption?}

In this paper, we address this question by proposing a simple yet effective preference optimization algorithm, named \method, which does not rely on the BT assumption. The key idea of \method is to directly optimize the conditional mutual information between responses and preferences given a prompt. We first revisit the DPO objective from the perspective of mutual information maximization. In particular, we demonstrate that the objective functions of DPO under the BT assumption can be characterized as special cases of a mutual information maximization objective, using InfoNCE~\cite{oord2018representation} as the estimator. Building on this insight, we propose a novel method that learns an effective policy from preference data without relying on the BT assumption. Specifically, \method leverages the NWJ estimator~\citep{nguyen2010estimating} for mutual information estimation instead of InfoNCE. Intuitively, \method weights the log likelihood of rejected responses according to model probability and uses an exponential operation to control the gradient norms towards rejected responses. We show that \method enables the model to update conservatively in the direction of rejected responses, while preventing a decrease in the likelihood of chosen responses. This results in improved downstream task performance, particularly in reasoning-heavy tasks.

We conduct extensive experiments to thoroughly evaluate \texttt{InfoPO} with LLama3 8B and Mistral 7B on a wide range of downstream benchmarks: Open LLM Leaderboard and AlpacaEval 2 and Anthropic-HH. \texttt{InfoPO} achieves consistent and significant improvements over existing  methods.

Our primary \textbf{technical contributions} are: \textbf{(1)} We propose a simple and effective alignment approach based on mutual information maximization, which can naturally prevent the model from over-fitting to preference data, striking a better balance between chat and reasoning abilities. \textbf{(2)} We theoretically analyze the learning
behavior and prove that \method enjoys some of the properties that
are desirable for fine-tuning  with preferences. \textbf{(3)} Empirically, we corroborate the effectiveness of \method
on seven benchmarks. The results demonstrate that \method can significantly outperform
previous preference optimization methods.

%% file: sections/related.tex
\section{Related Work}
\paragraph{Reinforcement Learning from Human Feedback.}
Reinforcement Learning from Human Feedback (RLHF) is highly effective in aligning Large Language Models (LLMs) with human preferences~\citep{ouyang2022training,christiano2017deep}. In RLHF, a reward model is trained from human preference data to map responses to a scalar reward, aligning a policy using RL algorithms such as PPO~\citep{schulman2017proximal}. Although RLHF excels in instruction-following~\citep{ouyang2022training}, safety alignment~\citep{bai2022training}, and summarization~\citep{stiennon2020learning}, RL fine-tuning for large language models still faces serious
challenges in stability and scalability~\citep{zheng2023secrets} and requires a more complex training pipeline compared to supervised fine-tuning (SFT) for alignment.

\paragraph{Contrastive Preference Fine-tuning.}
Recent work proposes simplifying RLHF by directly optimizing language models with contrastive learning on preference data~\citep{rafailov2024direct,azar2024general,ethayarajh2024kto,munos2023nash,liu2023statistical,xiao2024leverage,xiao2025simper}. While each of these methods work with different loss functions, the idea of them is to increase the gap between the likelihoods of preferred and dispreferred responses. DPO~\citep{rafailov2024direct} theoretically allows for direct policy optimization from preference data, equating its optimal solution to reward maximization in RLHF. Due to its strong performance and theoretical guarantees, several improvements have been proposed. RSO~\citep{liu2023statistical} uses rejection sampling to address sampling distribution mismatches in DPO, while IPO~\citep{azar2024general} prevents overfitting.
Other works~\citep{yuan2024self,xiong2023gibbs,rosset2024direct,guo2024direct} run DPO iteratively and on-policy. Despite these advances, the likelihood of the preferred response often decreases during DPO training, affecting performance on tasks such as coding and mathematics~\citep{pal2024smaug,yuan2024advancing}. 
Recent studies such as CPO~\cite{xu2024contrastive,pang2024iterative} propose using Negative Log Likelihood (NLL) regularization to stabilize training.
While these approaches successfully prevent collapse on mathematical datasets, they perform poorly on several popular Chat and QA benchmarks as shown in \cite{meng2024simpo}. In this paper, we address this limitation by proposing a new objective for alignment with preference data based on mutual information maximization. 

\paragraph{Mutual Information Estimation.}
Mutual information (MI) is a fundamental measure of the
dependence between two random variables. In machine learning, especially in deep learning frameworks, MI is typically utilized as a criterion or a regularizer in loss
functions, to encourage or limit the dependence between
variables. MI maximization has been studied extensively
in various tasks, e.g., representation learning~\cite{chen2020simple,bachman2019learning,tschannen2019mutual},
information distillation~\cite{ahn2019variational}, and reinforcement learning~\cite{oord2018representation,eysenbach2019diversity}. However, only in a few special cases can one calculate the exact value of mutual information, since the calculation
requires closed forms of density functions and a tractable
log-density ratio between the joint and marginal distributions. To approximate MI, there has been a surge of interest in MI estimation with variational approaches~\cite{barber2004algorithm,nguyen2010estimating,donsker1983asymptotic,belghazi2018mutual,oord2018representation,poole2019variational}. In this paper, we rethink the alignment on large language models from the mutual information maximization perspective.

\section{Notations and Preliminaries}

\textbf{Problem Setup.}
We consider the preference learning scenario as follows: let the text sequences $\mathbf{x} =[ x_1, x_2, \ldots ] \in X$ denote the input prompt, and $\mathbf{y}_{w}=[ y_1, y_2, \ldots ] \in Y$ and $\mathbf{y}_{l}=[ y_1, y_2, \ldots ] \in Y$ denote two responses, typically sampled from the reference policy $\pi_{\text{ref}}(\mathbf{y}|\mathbf{x})$. The response pairs are then presented to human labelers (or an oracle) who express preferences for responses given the prompt, denoted as $\mathbf{y}_{w} \succ \mathbf{y}_{l} | \mathbf{x}$, where $\mathbf{y}_{w}$ and $\mathbf{y}_{l}$ denote preferred and dispreferred responses, respectively. The preference distribution is typically expressed using a latent reward model $r(\mathbf{x},\mathbf{y})$ as:
\begin{align}
p\left(\mathbf{y}_w \succ \mathbf{y}_l | x\right)=g\left(r(\mathbf{x},\mathbf{y}_w)-r\left(\mathbf{x},\mathbf{y}_l\right)\right), \label{Eq:BT-model}
\end{align}
where $g: \mathbb{R} \rightarrow [0,1]$ is a monotone non-decreasing function (with $g(z) = 1 - g(-z)$) that converts reward differences into winning probabilities. When $g$ is the sigmoid function $\sigma(x) = \frac{1}{1+e^{-x}}$, we get the Bradley-Terry (BT) preference model~\citep{bradley1952rank}. Given dataset $\mathcal{D}$, containing feedback $(\mathbf{x}, \mathbf{y}_{w}, \mathbf{y}_{l})$, the goal is to learn a language model policy $\pi(\mathbf{y} \mid \mathbf{x})$ to align the human preference. 

\noindent \textbf{RLHF.} Given the reward function $r(\mathbf{x},\mathbf{y})$, denoting the human preferences, RLHF fine-tunes policy $\pi_\theta$  by
optimizing the following objective:
\begingroup\makeatletter\def\f@size{9.2}\check@mathfonts\def\maketag@@@#1{\hbox{\m@th\normalfont\normalfont#1}}
\begin{align}
\max_{{\boldsymbol{\theta}}} \mathbb{E}_{\pi_{\boldsymbol{\theta}}(\mathbf{y}|\mathbf{x})}\big[r(\mathbf{x},\mathbf{y})\big]- \beta \rm{KL}\big(\pi_{\boldsymbol{\theta}}(\mathbf{y}|\mathbf{x})\| \pi_{\rm{ref}}(\mathbf{y}|\mathbf{x})\big), \label{Eq:RL}
\end{align}
\endgroup
where $\beta > 0$ is an appropriate KL penalty coefficient. 
When $\beta \rightarrow 0$, all the probability mass will focus on the response with the highest reward. On the other extreme, when $\beta \rightarrow \infty$, the optimal policy will be the same as the reference policy $\pi_{\text{ref}}(\mathbf{y}|\mathbf{x})$. 
Due to the discrete nature of language generation, we typically optimize RLHF objective in Equation~\eqref{Eq:RL} using RL algorithms, such as PPO~\citep{ouyang2022training,schulman2017proximal}. Although RLHF with PPO has achieved remarkable success, the training process of PPO is unstable because of the high variance of the optimization~\citep{engstrom2020implementation,xiao2021general}.

\noindent \textbf{Reward Modeling.} One standard approach to reward modeling is to fit a reward function $r_{\phi}(\mathbf{x}, \mathbf{y})$  with the BT preference model in Equation~\ref{Eq:BT-model}. Specifically, the reward function $r_{\phi}(\mathbf{x}, \mathbf{y})$ can be estimated by maximizing the log-likelihood over preference feedback $(\mathbf{x},\mathbf{y}_{w},\mathbf{y}_{l})$:
\begin{align}
&\mathcal{L}_{\text{RM}}(\phi;\mathbf{x},\mathbf{y}_{w},\mathbf{y}_{l}) \nonumber \\
&=-\log \sigma \Big(r_{\phi}(\mathbf{x},\mathbf{y}_w)-r_{\phi}\left(\mathbf{x},\mathbf{y}_l\right)\Big).  \label{Eq:reward}
\end{align}

\noindent \textbf{DPO.} To simplify RLHF, contrastive preference learning ~\citep{tang2024generalized,rafailov2024direct,zhao2023slic,azar2024general} uses the log-likelihood of the learning policy to implicitly represent the reward function:
\begin{align}
r_\theta(\mathbf{x}, \mathbf{y})=\beta\left[\log \frac{\pi_{\theta}(\mathbf{y} | \mathbf{x})}{\pi_{\mathrm{ref}}(\mathbf{y}|\mathbf{x})} \right]+ \beta \log Z(\mathbf{x}),
\end{align}
where $Z(\mathbf{x})=\sum_{\mathbf{y}} \pi_{\text{ref}}(\mathbf{y} | \mathbf{x}) \exp(r(\mathbf{x},\mathbf{y})/\beta)$ is the partition function.
By incorporating this reward into the BT model in Equation~\ref{Eq:BT-model},  DPO~\citep{rafailov2024direct} objective enables the comparison of response pairs, facilitating the discrimination between preferred and dispreferred responses:
\begin{align}
&\mathcal{L}_{\rm{DPO}}(\theta;\mathbf{x},\mathbf{y}_{w},\mathbf{y}_{l})=  \label{Eq:DPO}\\
& -\log \sigma\Big(\beta \log \frac{\pi_{\theta}(\mathbf{y}_w | \mathbf{x})}{\pi_{\mathrm{ref}}(\mathbf{y}_w | \mathbf{x})}-\beta \log \frac{\pi_{\theta}(\mathbf{y}_l | \mathbf{x} )}{\pi_{\mathrm{ref}}(\mathbf{y}_l | \mathbf{x})}\Big). \nonumber
\end{align} 
Technically, DPO or its variants, such as those of SimPO~\citep{azar2024general} and ~\citep{zhao2023slic},  are essentially based on BT preference assumption~\citep{bradley1952rank} which maximizes the relative reward differences between chosen and rejected responses~\citep{tajwar2024preference}. However, the likelihood of the chosen response can continue to decrease during training as long as the relative difference in the likelihoods between the chosen and rejected responses remains large. In this paper, we address this limitation by proposing a novel objective based on mutual information maximization.

%% file: sections/method.tex
\section{Methodology}
\label{sec:method}
\subsection{Mutual Information Maximization for Large Language Alignment}
In this section, we connect \texttt{DPO} to mutual information maximization. We demonstrate that RLHF is a special case of the mutual information maximization problem by defining a specialized score function (or critic) approximated by a neural network. Specifically, we show that \texttt{DPO} can also be viewed as a special case of our framework by using contrastive predictive coding (\texttt{CPC}) (also known as \texttt{InfoNCE})~\cite{oord2018representation} for mutual information estimation. We focus on maximizing conditional mutual information~\cite{ma2021conditional}: $I(Y, C| X)$, where $C$ is an additional random variable. This variable is binary, with $\mathbf{c}=1$ indicating that the response is the preferred (chosen) one, and $\mathbf{c}=0$ indicating that it is the dispreferred (rejected) one. The conditional mutual information (CMI) is:
\begin{align}
&\text{CMI}(Y;C|X) :=  \label{Eq:CMT}\\
&\mathbb{E}_{\mathbf{x} \sim X} \left[\mathcal{D}_{\text{KL}} \left( P_{Y,C|X=\mathbf{x}} \parallel P_{Y|X=\mathbf{x}}P_{C|X=\mathbf{x}} \right) \right], \nonumber
\end{align}
which measures the expected mutual information between $C$ and $Y$ given $X$. Intuitively, $\text{CMI}(Y;C|X)$ quantifies the average shared information between $Y$ and $C$ while excluding the influence of $X$. Conditioning on $X = \mathbf{x}$ means treating $X = \mathbf{x}$ as known, thereby ignoring its effect. Since mutual information is often difficult to compute, \texttt{InfoNCE}~\cite{tsai2022conditional, ma2021conditional} provides a lower bound on the conditional mutual information as follows:
\begingroup\makeatletter\def\f@size{10}\check@mathfonts\def\maketag@@@#1{\hbox{\m@th\normalfont\normalfont#1}}
\begin{align}
&\text{CMI}(Y;C|X)\geq\text{InfoNCE}:= \\
&\sup_f \sum_{i=1}^{n} \Big [\log \frac{\exp ({f(\mathbf{y}_{i}, \mathbf{c}_{i}))}}{\exp({f(\mathbf{y}_{i}, \mathbf{c}_{i}))} + \sum_{j=1}^m {\exp(f(\mathbf{y}_{j}, \mathbf{c}_{j}))}} \Big],\nonumber
\end{align}
\endgroup
where the positive pairs, ${(\mathbf{y}_{i}, \mathbf{c}_{i})}_{i=1}^{n}$, represent samples drawn from the conditional joint distribution: $(\mathbf{y}_{i}, \mathbf{c}_{i}) \sim P_{Y,C|X}$, while the negative pairs, $(\mathbf{y}_{j}, \mathbf{c}_{j})$, represent samples drawn from the product of conditional marginal distributions: $(\mathbf{y}_{j}, \mathbf{c}_{j \neq i}) \sim P_{Y|X} P_{C|X}$. The score function $f$ can be approximated by a neural network. Given the prompt distribution $p(\mathbf{x})$ and the conditional distribution of the preferred response $\pi(\mathbf{y}, \mathbf{c=1} \mid \mathbf{x})$, we sample $\mathbf{x} \sim p(\mathbf{x})$, $(\mathbf{y}_{w},\mathbf{c}) \sim \pi(\mathbf{y}, \mathbf{c=1} \mid \mathbf{x})$, and $(\mathbf{y}_{l},\mathbf{c}) \sim \pi_{\rm{ref}}(\mathbf{y} \mid \mathbf{x})p(\mathbf{c} \mid \mathbf{x})$. The objective of \texttt{InfoNCE} with preference feedback is as follows:
\begin{align}
&\mathcal{L}_{\rm{InfoNCE}}(\boldsymbol{\phi};\mathbf{x},\mathbf{y}_{w},\mathbf{y}_{l}) =   \\
&-\log \frac{\exp(f_{\boldsymbol{\phi}}(\mathbf{y}_{w},\mathbf{c}=1 ))}{\exp (f_{\boldsymbol{\phi}}(\mathbf{y}_{w},\mathbf{c}=1 )) +\exp(f_{\boldsymbol{\phi}}(\mathbf{y}_{l},\mathbf{c}=0 ))}, \nonumber
\end{align}
where $f_{\boldsymbol{\phi}}$ is a parametric critic function.
If we define the critic with the following specialized form: 
\begin{align}
f_{\boldsymbol{\phi}}(\mathbf{x}, \mathbf{c})=\beta \log \frac{\pi_{\boldsymbol{\theta}}(\mathbf{y} \mid \mathbf{x})}{\pi_{\rm{ref}}(\mathbf{y}\mid \mathbf{x})}, \label{Eq:critic}
\end{align}
we have the following \texttt{InfoNCE} objective function: 
\begin{align}
&\mathcal{L}_{\text{InfoNCE}}(\boldsymbol{\theta};\mathbf{x},\mathbf{y}_{w},\mathbf{y}_{l})= \\
&-  \log \sigma \Big( \beta \log \frac{\pi_{\boldsymbol{\theta}}(\mathbf{y}_{w} \mid \mathbf{x})}{\pi_{\text{ref}}(\mathbf{y}_{w} \mid \mathbf{x})} - \beta \log \frac{\pi_{\boldsymbol{\theta}}(\mathbf{y}_{l} \mid \mathbf{x})}{\pi_{\text{ref}}(\mathbf{y}_{l} \mid \mathbf{x})} \Big) , \nonumber
\end{align}
which is exactly the same objective as the well-known \texttt{DPO} in Equation~\eqref{Eq:DPO}. Thus, our framework enables us to reinterpret \texttt{DPO} and  demonstrate that \texttt{DPO} with BT assumption falls under the conditional mutual information maximization $I(Y;C|X)$ in Equation~(\ref{Eq:CMT}) and  employs the \texttt{InfoNCE} method with a specialized form critic in Equation~\eqref{Eq:critic}.

\subsection{Gradient Analysis of \texttt{DPO}}
To better understand the reason behind the behavioral of \texttt{DPO} in optimization, we analyze the gradients of \texttt{DPO} with respect to the model parameters.
\begin{align}
&\nabla_{\theta} \mathcal{L}_{\text{{DPO}}}({\theta};\mathbf{x},\mathbf{y}_{w},\mathbf{y}_{l})= \label{Eq:DPO-gradient} \\
& -\beta d_{\theta} \cdot \Big( \frac{\nabla_{\theta} \pi_{\theta}(\mathbf{y}_w| \mathbf{x})}{\pi_{\theta}(\mathbf{y}_w| \mathbf{x})} - \frac{\nabla_{\theta} \pi_{\theta}(\mathbf{y}_l | \mathbf{x})}{\pi_{\theta}(\mathbf{y}_l | \mathbf{x})}  \Big),  \nonumber 
\end{align} 
where $d_{\theta} = \sigma ( \beta \log \frac{\pi_{\theta}(\mathbf{y}_l | \mathbf{x})}{\pi_{\text{ref}}(\mathbf{y}_l | \mathbf{x})} - \beta \log \frac{\pi_{\theta}(\mathbf{y}_w| \mathbf{x})}{\pi_{\text{ref}}(\mathbf{y}_w| \mathbf{x})})$ represent the gradient weight in \texttt{DPO}. It can be observed that the gradient of the model probability, weighted by the reciprocal of the model probability, is large for the rejected response. Intuitively, the gradient of \texttt{DPO} increases the likelihood of the chosen response, $\mathbf{y}_{w}$, while decreasing the likelihood of the rejected response, $\mathbf{y}{l}$. If $\pi_{\theta}(\mathbf{y}_{l}|\mathbf{x}) \rightarrow 0 $, the norm of the gradient becomes extremely large, leading to a substantial parameter update toward the rejected response. In this scenario, the gradient associated with the rejected response grows excessively large, whereas the gradient for the chosen response diminishes significantly. This explains the phenomenon illustrated in Figure~\ref{fig:rewards-mistral} in the introduction, where $\pi_{\theta}$ forces the model to decrease the likelihood of the chosen response during training, given that the rejected and chosen responses share some common tokens~\citep{pal2024smaug,meng2024simpo,xiaocal}.

\subsection{The Proposed  \method}
\label{sec:density}
In this section, we proceed to introduce \texttt{InfoPO}, a simple and effective preference optimization algorithm. Instead of using \texttt{InfoNCE} for mutual information estimation in \texttt{DPO}, we propose using the following NWJ~\cite{nguyen2010estimating} estimator:
\begingroup\makeatletter\def\f@size{10}\check@mathfonts\def\maketag@@@#1{\hbox{\m@th\normalfont\normalfont#1}}
\begin{align}
&\text{CMI}(Y;C|X)\geq\text{NWJ}:= \nonumber \\
&\sup_f \sum_{i=1}^{n} {{f(\mathbf{y}_{i}, \mathbf{c}_{i})}}-\sum_{j=1}^{m}{\exp\big({f(\mathbf{y}_{j}, \mathbf{c}_{j})\big)} }+1. \label{Eq:NWJ2}
\end{align}
\endgroup
By using the preference datasets and the parameterized critic in Equation~\eqref{Eq:critic}, we have the following \texttt{InfoPO} objective on preference pairs:
\begin{align}
&\mathcal{L}_{\rm{InfoPO}}({\theta};\mathbf{x},\mathbf{y}_{w},\mathbf{y}_{l})= \label{Eq:InfoPO} \\
&- \log \pi_{\theta} (\mathbf{y}_{w}|\mathbf{x})+ {\pi_{\theta} (\mathbf{y}_{l}|\mathbf{x})}/{\pi_{\rm{ref}} (\mathbf{y}_{l}|\mathbf{x})}. \nonumber
\end{align}
Intuitively, the first term pushes the model to minimize the negative log-likelihood (NLL) of the chosen response, while the second term decreases the likelihood of the rejected response. The key contribution behind \texttt{InfoPO} is rather simple yet effective: If the gradients of both chosen and rejected responses lie on the same scale, we can prevent the reward (likelihood) of chosen responses from continually decreasing. \texttt{InfoPO} utilizes an exponential operation  to linearize gradients on rejected responses. For comparison, we calculate the gradient of \texttt{InfoPO} with respect to $\theta$ using Equation~(\ref{Eq:InfoPO}): 
\begin{align}
    \nabla_{\theta} \mathcal{L}&_{\rm{InfoPO}}({\theta};\mathbf{x},\mathbf{y}_{w},\mathbf{y}_{l})= \nonumber  \\
   & -\frac{\nabla_{\theta}\pi_{\theta}(\mathbf{y}_w| \mathbf{x})}{\pi_{\theta}(\mathbf{y}_w| \mathbf{x})} +\frac{\nabla_{\theta} \pi_{\theta}(\mathbf{y}_l | \mathbf{x})}{\pi_{\rm{ref}}(\mathbf{y}_l | \mathbf{x})}
\end{align}
where the gradient weight of rejected response is the reciprocal of the fixed reference probability of the sample, which has a smaller norm than Equation~\eqref{Eq:DPO-gradient}. This means that the  unlearning on the rejected responses is more conservative and \texttt{InfoPO} reduces the gradient imbalance issue for chosen and rejected responses. In our experiment, we show that \texttt{InfoPO} can effectively prevent the chosen likelihood from decreasing and significantly outperforms baselines across benchmarks. NWJ (InfoPO) and InfoNCE (DPO) exhibit different properties for conditional mutual information estimation, and their performance varies depending on the specific scenario. Specifically, for mutual information maximization, NWJ has low bias but high variance, whereas InfoNCE has low variance but suffers from high bias, as shown in~\cite{poole2019variational}.

\subsection{Theoretical Analysis}
Next, we proceed to present a theoretical analysis of \texttt{InfoPO}, which shows that \texttt{InfoPO}  enjoys important properties that are desirable  for fine-tuning LLMs with preferences~\cite{tajwar2024preference}.
\begin{theorem}
\label{the:mutual}
Minimizing the \text{\texttt{InfoPO}} objective in Equation~(\ref{Eq:InfoPO}) with respect to ${\theta}$ will encourage mode-seeking behavior by minimizing the reverse KL divergence between  $\pi_\theta (\mathbf{y}\mid\mathbf{x})$ and unknown distribution of chosen response $\pi_{\rm{chosen}}(\mathbf{y}\mid\mathbf{x})$.
\begingroup\makeatletter\def\f@size{9.5}\check@mathfonts\def\maketag@@@#1{\hbox{\m@th\normalfont\normalfont#1}}
\begin{align}
&\min_{\theta} \mathcal{L}_{\rm{InfoPO}}({\theta}) \Rightarrow \min_{\theta}\mathcal{D}_{\mathrm{KL}}(\pi_\mathbf{\theta}(\mathbf{y}\mid\mathbf{x}) \| \pi_{\rm{chosen}}(\mathbf{y}\mid\mathbf{x}) ) \nonumber \\
&=\mathbb{E}_{\pi_\mathbf{\theta}(\mathbf{y}\mid\mathbf{x})}\left[\log {\pi_\mathbf{\theta}(\mathbf{y}\mid\mathbf{x})}-\log {\pi_{\rm{chosen}}(\mathbf{y}\mid\mathbf{x})} \right].
\end{align}
\endgroup
\end{theorem}
\noindent The complete proof is provided in Appendix~\ref{app:prove}.
This theorem demonstrates that \texttt{InfoPO} theoretically minimizes the reverse KL divergence between the policy $\pi_\theta$ and the unknown distribution of the chosen response $\pi_{\rm{chosen}}$. 
The reverse ${\mathrm{KL}}(\pi_{\boldsymbol{\theta}} \parallel \pi_{\rm{chosen}})$ promotes mode-seeking behavior, concentrating the probability mass on high-reward regions. This makes reverse KL more suitable for alignment aimed at generating a focused subset of high-reward responses, as demonstrated by~\cite{tajwar2024preference,xiaocal}.

%% file: sections/experiments.tex
\section{Experiments}
\label{sec:exp}
In this section, we present main results of our experiments, highlighting the superior alignment performance of
\texttt{InfoPO} on various benchmarks.

\begin{table*}[t]
    \centering
    \caption{ Evaluation results on tasks from the Huggingface Open Leaderboard and AlpacaEval 2.}
    \setlength\tabcolsep{5pt}
    \label{table:metric_comparison}
    \scalebox{0.95}{
    \begin{tabular}{ll*{7}{c}}
    \toprule[1pt]
  \textbf{Model} & \multicolumn{1}{c}{\textbf{Method}}  
    & \textbf{MMLU-PRO} 
    & \textbf{BBH} 
    & \textbf{MUSR} 
    & \textbf{MATH} 
    & \textbf{GSM8K} 
    & \textbf{ARC} 
    & \textbf{AlpacaEval 2}  \\
    \midrule[0.5pt]
    \multirow{8}{*}{{\centering Mistral-7B}}
    & DPO 
    & 26.73 
    & {43.27} 
    & {43.65} 
    & 1.36 
    & 21.76 
    & 61.26  
    & 12.5   \\
    & SLiC 
    & 26.52 
    & 42.33 
    & {33.74} 
    & 1.38 
    & \textbf{33.74}  
    & 55.38  
    & 8.9\\
    & f-DPO  
    & 25.96 
    & 42.39 
    & 37.82 
    & 1.27 
    & 23.18 
    & 62.01 
    & 8.5  \\
    & IPO 
    & 25.87 
    & 40.59
    & 42.15 
    & 1.25 
    & 27.14  
    & 60.84  
    & 9.4 \\
    & CPO 
    & 27.04 
    & 42.05 
    & 42.15 
    & 2.15 
    & {33.06} 
    & 57.00 
    & 8.9 \\
    & SimPO
    & {27.13} 
    & 42.94 
    & 39.68 
    & {2.49} 
    & 22.21 
    & \textbf{62.63} 
    & {20.8}  \\
    \cmidrule(lr){2-9}
    & InfoPO
    & \textbf{27.32}
    & \textbf{45.17}
    & \textbf{43.95}
    & \textbf{2.79}
    &  32.07
    & {62.29}
    & \textbf{21.6} \\
    \midrule[0.8pt]
    \multirow{8}{*}{{\centering LLama3-8B}}
    & DPO 
    & 31.58 
    & 47.80 
    & 40.48
    & {4.53} 
    & 38.67  
    & 64.42 
    & 15.5  \\
    & SLiC 
    & 31.11 
    & 46.53 
    & 40.55 
    & 3.92 
    & {48.82} 
    & 61.43 
    & 13.7\\
    & f-DPO 
    & 30.85 
    & 47.55 
    & 40.39 
    & 4.37 
    & 39.55 
    & 62.85 
    & 9.5 \\
    & IPO 
    & 30.18 
    & 46.78 
    & 39.58 
    & 4.02 
    & 22.67 
    & 62.88 
    & 14.2  \\
    & CPO
    & 30.95 
    & 47.17 
    & {41.59} 
    & 4.25 
    & 46.93 
    & 61.69 
    & 8.10 \\
    & SimPO
    & {31.61} 
    & {48.38} 
    & 40.08 
    & 4.23 
    & 31.54 
    & {65.19} 
    & {20.3} \\
    \cmidrule(lr){2-9}
    & InfoPO
    & \textbf{32.06}
    & \textbf{48.85}
    & \textbf{42.31}
    &  \textbf{4.69}
    & \textbf{49.13}
    & \textbf{65.36}
    & \textbf{26.6} \\
    \bottomrule[1pt]
    \end{tabular}
    }
\end{table*}

\subsection{Experimental Setup}

\paragraph{Datasets.} We evaluate \method ~on widely used datasets for preference fine-tuning:  UltraFeedback Binarized dataset~\citep{cui2023ultrafeedback},  Reddit TL;DR summarization dataset~\citep{volske2017tl},  Anthropic-HH dialogue  dataset~\citep{bai2022training}. The details of datasets are given in Appendix~\ref{app:dataset}.

\paragraph{Models.} 
For fine-tuning on the UltraFeedback Binarized dataset, we use two families of models: Llama3-8B~\citep{dubey2024llama} and Mistral-7B~\citep{jiang2023mistral}, following~\citep{meng2024simpo}. For summarization and dialogue generation tasks, we use Pythia-2.8B~\cite{biderman2023pythia} as the base model, following~\cite{rafailov2024direct}.

\begin{figure*}[t]
\centering 
\includegraphics[width=0.999\textwidth]{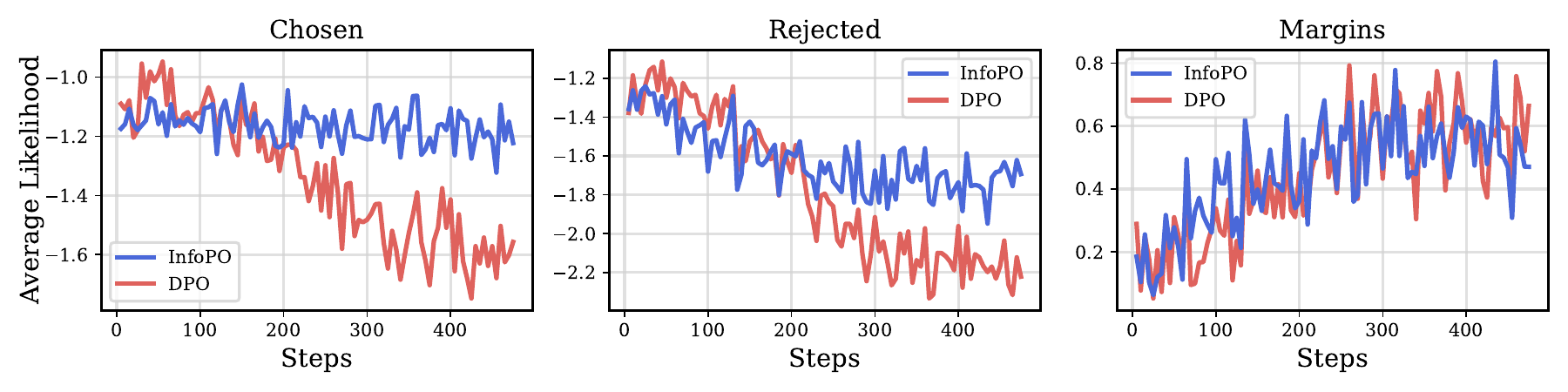}
\vskip -0.5em
\caption{The training dynamics of average likelihood  of \method and \texttt{DPO} on the Llama3-8B. We observe that \method exhibits the less decline in the average chosen likelihoods, while still achieving the significant increase in  margins of rejected and chosen likelihood, compared to \texttt{DPO}.}
\vskip -0.5em
\label{fig:rewards-llama} 
\end{figure*}

\begin{table*}[t]
\centering
\caption{Win rates computed by GPT-4 against the   response generated by the model with supervised fine-tuning and the chosen responses on the TL;DR summarization and Anthropic-HH dialogue tasks on Pythia 2.8B.}\label{tab:preference}
    \setlength\tabcolsep{12pt}
\scalebox{0.91}{
\begin{tabular}{@{}cc>{}c>{}cc>{}c>{}c@{}}
\toprule
\textbf{Dataset} & \multicolumn{3}{c}{\textbf{TL;DR Summarization}} & \multicolumn{3}{c}{\textbf{Anthropic-HH Dialogue}} \\
\cmidrule(lr){2-4} \cmidrule(lr){5-7}
\textbf{Method}  & \textbf{vs SFT} & {\textbf{vs Chosen}} & {\textbf{Average} }& \textbf{vs SFT} & {\textbf{vs Chosen}} & {\textbf{Average} }\\ \midrule
DPO  & 71.22 &  57.58    & 64.40  &   69.32   & 59.35   & 64.34  \\
SLiC  & 68.61 &  55.72    & 62.17  &   65.52   & 57.71   & 61.62  \\
f-DPO  & 66.19 &  51.37    & 58.78  &   60.21  &  52.38   & 56.30  \\
IPO & 72.17 & 56.51 & 64.34 &   63.19  &  55.12   & 59.16  \\
CPO & \underline{{73.13}}  & \underline{{58.89}} & \underline{{66.01}} &   \underline{{72.30}}  &  \underline{{63.39}}   & \underline{{67.86}} \\
SimPO & 69.71  & 54.38 & 62.05  &   67.85  & 57.51  & 62.68 \\
\midrule
InfoPO
&  \textbf{73.95}
& \textbf{60.12}
& \textbf{67.04}
&  \textbf{73.38} 
&  \textbf{64.85}  
&   \textbf{69.12}   \\
\bottomrule
\end{tabular}}
\end{table*}

\paragraph{Evaluation.} Following previous work~\citep{rafailov2024direct,tunstall2023zephyr}, we evaluate methods fine-tuned on UltraFeedback Binarized  on the HuggingFace Open LLM Leaderboard~\citep{eval-harness} and instruction-following benchmark (AlpacaEval2).  We also utilize representative code generation benchmarks: HumanEval~\citep{chen2021evaluating} and MBPP~\citep{austin2021program}. For the evaluation on summarization and dialogue generation tasks, we use GPT-4 for zero-shot pair-wise evaluation following~\cite{rafailov2024direct}, which is shown to be consistent with human judgments.

\paragraph{Baselines.} We compare \texttt{InfoPO} with following  offline preference optimization methods: DPO~\citep{rafailov2024direct}, f-DPO~\citep{DBLP:conf/iclr/WangJYLC24},  IPO~\citep{azar2024general}, and SimPO~\citep{meng2024simpo}.  We also compare with CPO~\cite{xu2024contrastive}, which is a representative method of introducing a SFT regularization to prevent the decrease of chosen likelihood.  We thoroughly tuned the hyperparameters for each baseline. For the general hyperparameter settings, we follow the configurations established in SimPO~\cite{meng2024simpo}. The details of   setup is given in Appendix~\ref{app:setup}.

\subsection{Main Results on Benchmarks}
We first employ the widely used Huggingface Open LLM Leaderboard and AlpacaEval 2 as our evaluation benchmarks. Table~\ref{table:metric_comparison} compares the performance of \texttt{InfoPO} against other preference optimization methods. Our results demonstrate that \texttt{InfoPO} is remarkably effective in improving performance. The average improvements over the best baseline are particularly notable in the Math domain, with relative gains exceeding 12\% on Mistral and 3.5\% on Llama3. These findings highlight the efficacy of \texttt{InfoPO}. We hypothesize that these improvements can be attributed to \texttt{InfoPO}'s ability to prevent decreases in the chosen response during training. Additionally, the results suggest that DPO and SimPO are less effective for enhancing reasoning abilities, while \texttt{InfoPO} shows clear improvements on both the Mistral-7B and Llama3-8B.

In addition to the reasoning tasks, we also compare the performance of \texttt{InfoPO} on the instruction-following benchmark, AlpacaEval 2. The win rate results on AlpacaEval 2 in Table~\ref{table:metric_comparison} demonstrate that \texttt{InfoPO} consistently and significantly outperforms existing alignment approaches. We further evaluate the model's performance on coding tasks using HumanEval~\cite{chen2021evaluating} and MBPP~\cite{austin2021program}. The results, presented in Table~\ref{table:metric_comparison}, show that \texttt{InfoPO} consistently outperforms both DPO and SimPO. This further indicates that \texttt{InfoPO} is more suitable for enhancing reasoning abilities compared to DPO and SimPO.

\subsection{Performance Comparisons on Summarization and Dialogue Tasks}
We also assess the performance of \texttt{InfoPO} on summarization and dialogue generation tasks. As shown in Table~\ref{tab:preference}, \texttt{InfoPO} demonstrates substantial improvements over the baseline models in both tasks. These results highlight \texttt{InfoPO}'s capability to enhance not only reasoning and coding abilities but also natural language generation in diverse applications. Specifically, \method ~aligns better with human preferences than baselines, achieving a win rate of at least 60\% against the chosen responses in both tasks. This highlights the strong potential of \method ~for aligning with human preferences. Furthermore, GPT-4 consistently favored \method ~over both baselines and chosen responses, demonstrating improvements of \method over baselines in both helpfulness and harmlessness. The superior performance of \texttt{InfoPO} on these tasks further emphasizes its effectiveness in multiple domains.

\subsection{Further Results and Analysis}
\label{exp:anly}
In this subsection, we take a deeper examination and further analysis on the proposed
framework.
\subsubsection{Performance on  On-Policy Settings.}
In the above experiments, we utilize the offline preferences dataset to fine-tune the off-the-shelf language models, which is closer to an off-policy setting. To evaluate \method on the on-policy setting, we follow the instruct setup in~\cite{meng2024simpo}. We generate the preference dataset using the LLama3-Instruct~\cite{dubey2024llama} and  Mistral-Instruct~\cite{jiang2023mistral} models.  Specifically, we use prompts from the UltraFeedback dataset and regenerate the chosen and rejected response pairs  with the SFT models. For each prompt $\mathbf{x}$, we generate $5$ responses using the SFT model with a sampling temperature of $0.8$. We then use llm-blender/PairRM~\citep{jiang2023llm} to score the five responses, selecting the highest-scoring one as the chosen response and the lowest-scoring one as the rejected response. This makes the Instruct setup closer to an on-policy setting. Table~\ref{table:on-policy} shows the results. From the table, we can observe that, \method, despite its simplicity, achieves remarkable improvements over DPO, CPO, and SimPO, particularly on challenging reasoning benchmarks such as Math and GSM8K, demonstrating that \method is highly effective in improving reasoning performance over various models on the on-policy scenario.

\begin{figure}
\centering
\includegraphics[width=0.475\textwidth]{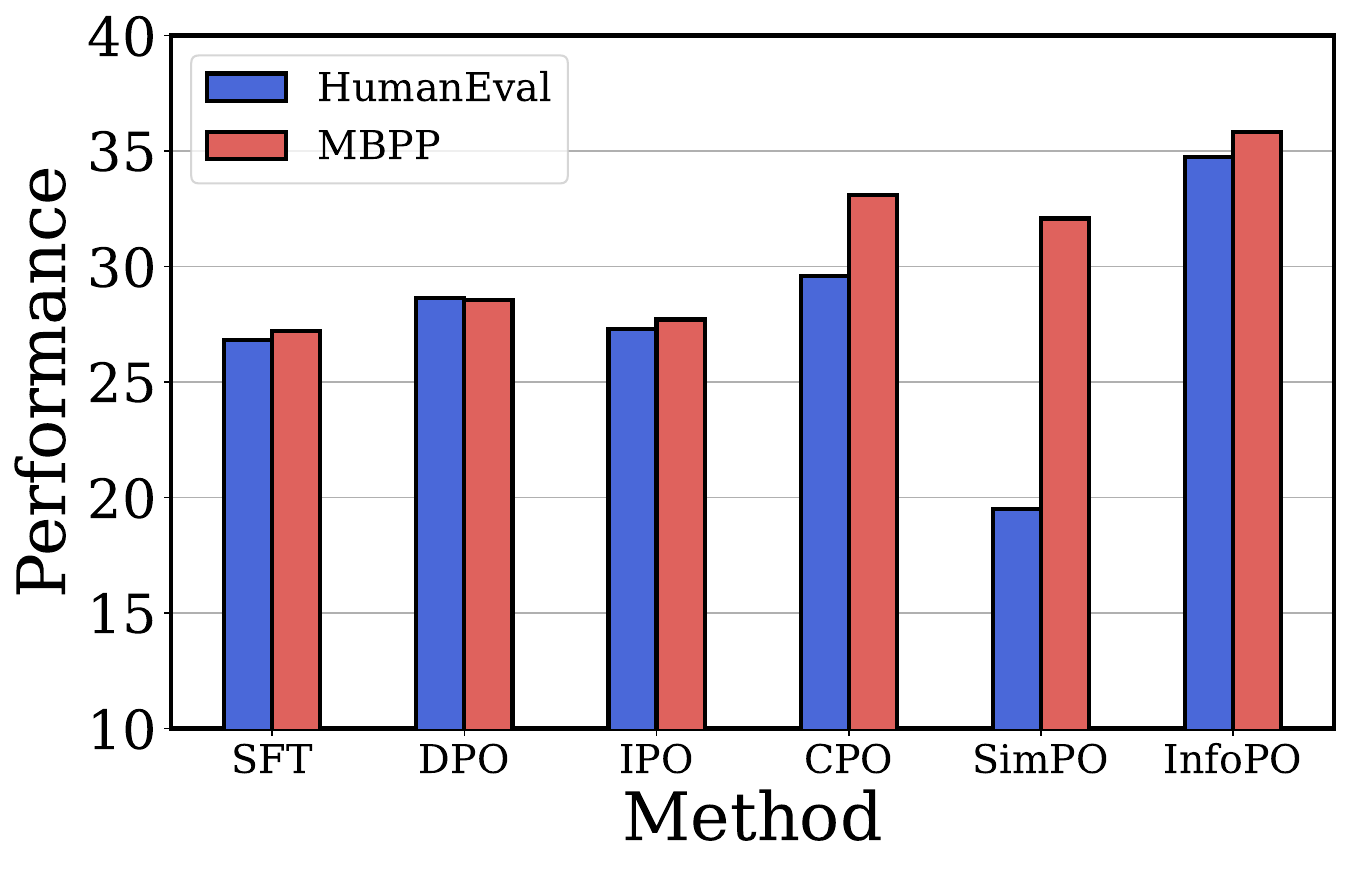}
\vskip -0.5em
     \caption{The performance comparison on coding tasks. }
\label{fig:coding} 
\vskip -0.5em
\end{figure}
\subsubsection{Likelihood Training Dynamics.}
We further examine the behavior of likelihoods throughout the training process of \method. As illustrated in Figure \ref{fig:rewards-llama}, we compare the likelihood trajectories of \texttt{DPO} and \method on the Llama3-8B model. It is evident that the likelihood of rejected responses consistently declines, with the gap between chosen and rejected responses widening over time. However, for \texttt{DPO}, the likelihood of chosen responses not only drops below zero but continues to decrease as training progresses. This outcome reinforces our hypothesis, highlighting \method’s ability to prevent a decline in the likelihood of chosen responses. This likely contributes to the superior performance of \method on downstream tasks, particularly those involving complex reasoning, such as math and coding, as demonstrated in results in Table~\ref{table:metric_comparison} and Figure~\ref{fig:coding}.

\section{Conclusion}
This paper presents \texttt{InfoPO}, a novel preference fine-tuning method  to align LLMs with  preference data. We provide a novel perspective on mutual information maximization for the alignment problem, and demonstrate DPO with BT assumption essentially optimize  the  contrastive \texttt{InfoNCE} objective. To address the limitation of DPO, we propose  \texttt{InfoPO}  based on \texttt{NWJ} mutual information estimator.  \texttt{InfoPO}  applies an exponential function to control gradient magnitudes associated with these rejected responses.  \texttt{InfoPO} enables the model to update more conservatively in response to rejections, thereby reducing the likelihood of overestimating such responses. We have conducted a comprehensive evaluation of \texttt{InfoPO} on different LLMs across a broad downstream tasks. Experimental results show that \texttt{InfoPO} achieves consistent and substantial improvements over existing baselines.

\setlength{\tabcolsep}{1pt}
\begin{table}
    \centering
  \fontsize{9}{10}\selectfont\setlength{\tabcolsep}{0.5em}
    \caption{On-policy evaluation results on reasoning tasks (GSM8K and Math) in  Huggingface Open Leaderboard. }
\label{table:on-policy}
    \resizebox{0.5\textwidth}{!}{%
    \begin{tabular}{ll*{9}{c}}
    \toprule[1pt]
    \multicolumn{1}{c}{\textbf{Model}}  
    & \textbf{Method} 
    & \textbf{MUSR} 
    & \textbf{MATH} 
    & \textbf{GSM8K}  \\
    \midrule[0.5pt]
    \multirow{4}{*}{\parbox[t]{1.6cm}{\centering Mistral-7B \\ Instruct}}
    & DPO
    & 46.43
    & 1.89
    & 35.25  \\
    & CPO
    & 43.28
    & 2.28
    & 38.74  \\
    & SimPO
    & 44.71
    & 2.19
    & 35.25 \\
    \cmidrule(lr){2-8}
    & InfoPO
    &  \textbf{48.41}
    & \textbf{2.64} 
    & \textbf{40.87} \\
    \midrule[0.7pt] 
    \multirow{4}{*}{\parbox[t]{1.6cm}{\centering LLama3-8B \\ Instruct}}
    & DPO
    & 39.02
    & 8.23
    &  49.81  \\
    & CPO
    &  38.81
    & 7.75
    & {67.40}  \\
    & SimPO
    & 39.15
    & 8.16
    & {50.72}  \\
    \cmidrule(lr){2-8}
    & InfoPO
    & \textbf{39.37}
    & \textbf{8.79}
    & \textbf{69.75} \\
    \bottomrule[1pt]
\end{tabular}}
\end{table}

\section{Limitations and Future Work}
One limitation of \method is its current reliance on a single mutual information estimator. While this work primarily employs the NWJ mutual information estimation loss function, exploring the effectiveness of \method with alternative mutual information estimators remains an interesting avenue for future research. Additionally, gaining a deeper theoretical understanding of which mutual information estimation techniques are most effective for alignment is a key direction for further study.

%% file: sections/appendix-proof.tex
\newpage
\appendix
\section{Proof of Theorem~\ref{the:mutual}}
\label{app:prove}
In this section, we provide the detailed proofs of Theorem~\ref{the:mutual}.  Here, we restate the Theorem~\ref{the:mutual}.

\noindent  \textit{ \textbf{Theorem 4.1}
Minimizing the \text{\texttt{InfoPO}} objective in Equation~(\ref{Eq:InfoPO}) with respect to ${\theta}$ will encourage mode-seeking behavior by minimizing the reverse KL divergence between  $\pi_\theta (\mathbf{y}\mid\mathbf{x})$ and unknown distribution of chosen response $\pi_{\rm{chosen}}(\mathbf{y}\mid\mathbf{x})$.}
\begingroup\makeatletter\def\f@size{9.5}\check@mathfonts\def\maketag@@@#1{\hbox{\m@th\normalfont\normalfont#1}}
\begin{align}
&\min_{\theta} \mathcal{L}_{\rm{InfoPO}}({\theta}) \Rightarrow \min_{\theta}\mathcal{D}_{\mathrm{KL}}(\pi_\mathbf{\theta}(\mathbf{y}\mid\mathbf{x}) \| \pi_{\rm{chosen}}(\mathbf{y}\mid\mathbf{x}) ) \nonumber \\
&=\mathbb{E}_{\pi_\mathbf{\theta}(\mathbf{y}\mid\mathbf{x})}\left[\log {\pi_\mathbf{\theta}(\mathbf{y}\mid\mathbf{x})}-\log {\pi_{\rm{chosen}}(\mathbf{y}\mid\mathbf{x})} \right].
\end{align}
\endgroup
\begin{proof}
Recall that the reverse KL-divergence between $\pi_{\boldsymbol{\theta}}$ and the chosen distribution ~$\pi_{\rm{chosen}}$ is: 
\begingroup\makeatletter\def\f@size{9.5}\check@mathfonts\def\maketag@@@#1{\hbox{\m@th\normalfont\normalfont#1}}
\begin{align}
&\mathcal{D}_{\mathrm{KL}}\Big(\pi_{\boldsymbol{\theta}}(\mathbf{y}\mid \mathbf{x})\| \pi_{\rm{chosen}}(\mathbf{y}\mid \mathbf{x})\Big) \nonumber \\ &= \mathbb{E}_{\pi_{\boldsymbol{\theta}}(\mathbf{y} \mid \mathbf{x})} \Big[\log \Big({\pi_{\boldsymbol{\theta}}(\mathbf{y}\mid\mathbf{x})}/{\pi_{\rm{chosen}}(\mathbf{y}\mid \mathbf{x})}\Big)\Big], \label{Eq:IL}
\end{align}
\endgroup
As the chosen  distribution is unknown,  we reformulate the reverse KL divergence objective as:
\begingroup\makeatletter\def\f@size{9}\check@mathfonts\def\maketag@@@#1{\hbox{\m@th\normalfont\normalfont#1}}
\begin{align}
&\max_{{\boldsymbol{\theta}}}\mathbb{E}_{\pi_{\boldsymbol{\theta}}(\mathbf{y}\mid\mathbf{x})}\Big[\log \frac{\pi_{\rm{chosen}}(\mathbf{y}\mid\mathbf{x})}{\pi_{\rm{ref}}(\mathbf{y}\mid \mathbf{x})}-\log \frac{\pi_{\boldsymbol{\theta}}(\mathbf{y}\mid \mathbf{x})}{\pi_{\rm{ref}}(\mathbf{y}\mid \mathbf{x})}\Big]= \nonumber  \\
&\mathbb{E}_{\pi_{\boldsymbol{\theta}}(\mathbf{y}|\mathbf{x})}[\log r(\mathbf{x},\mathbf{y})]-{\rm{KL}}\big(\pi_{\boldsymbol{\theta}}(\mathbf{y}\mid \mathbf{x})\| \pi_{\rm{ref}}(\mathbf{y}\mid \mathbf{x})\big),\label{Eq:surrogate}
\end{align}
\endgroup
where $r(\mathbf{x}, \mathbf{y}) \triangleq \frac{\pi_{\rm{chosen}}(\mathbf{y} \mid \mathbf{x})}{\pi_{\rm{ref}}(\mathbf{y} \mid \mathbf{x})}$ can be viewed as an auxiliary reward function. Equations (\ref{Eq:IL}) and (\ref{Eq:surrogate}) are equivalent by adding and subtracting the same term of $\log {\pi_{\rm{ref}}(\mathbf{y}\mid \mathbf{x})}$ in the expectation. 
In the tabular setting, we can directly compute $\pi_{\rm{ref}}(\mathbf{y}\mid\mathbf{x})$ and $\pi_{\rm{chosen}}(\mathbf{y}\mid\mathbf{x})$. However, in a high-dimensional language domain, estimating the densities separately and then calculating their ratio hardly works well due to error accumulation. However, we can directly estimate the density ratio ${\pi_{\rm{chosen}}(\mathbf{y} \mid \mathbf{x})}/{\pi_{\rm{ref}}(\mathbf{y} \mid \mathbf{x})}$ based on mutual information.  A simple alternative 
is to estimate the log ratio via learning a discriminator with the
following NWJ~\cite{nguyen2010estimating} estimator:
\begingroup\makeatletter\def\f@size{10}\check@mathfonts\def\maketag@@@#1{\hbox{\m@th\normalfont\normalfont#1}}
\begin{align}
\mathbb{E}_{\pi_\mathrm{ref}}\left[\exp \left(f(\mathbf{x},\mathbf{y})\right)\right] - \mathbb{E}_{\pi_\mathrm{chosen}} \left[\big(f(\mathbf{x},\mathbf{y})\big)\right], \label{Eq:KLIEF}
\end{align}
\endgroup
The log density ratio are related to the optimal discriminator~\cite{songunderstanding}:
\begin{align}
    f^{*}(\mathbf{x},\mathbf{y})=\log \frac{\pi_\mathrm{chosen}(\mathbf{y}\mid \mathbf{x})}{\pi_{\rm{ref}}(\mathbf{y}\mid \mathbf{x})}.
\end{align}
Thus the RL-style objective in Equation~(\ref{Eq:surrogate}), combined with density ratio estimation in Equation~(\ref{Eq:KLIEF}), can effectively optimize the reverse KL divergence. 
we can  directly optimize the reverse KL divergence, bypassing the need for RL training and density ratio estimation. The key idea is to leverage a specific discriminator parameterization, enabling a direct extraction of optimal policy, without an RL loop. Specifically, the optimal policy in~\eqref{Eq:surrogate} has a closed form~\cite{rafailov2024direct}:
\begingroup\makeatletter\def\f@size{9.5}\check@mathfonts\def\maketag@@@#1{\hbox{\m@th\normalfont\normalfont#1}}
\begin{align}
\pi^{*}(\mathbf{y} \mid \mathbf{x})=\frac{1}{Z(\mathbf{x})} \pi_{\rm{ref}}(\mathbf{y} \mid \mathbf{x}) \exp \left( f^{*}(\mathbf{x}, \mathbf{y})\right), \label{Eq:optimal}
\end{align}
\endgroup
where $Z(\mathbf{x})=\sum_{\mathbf{y}} \pi_{\rm{ref}}(\mathbf{y}| \mathbf{x}) \exp \left( r(\mathbf{x}, \mathbf{y})\right)=\sum_{\mathbf{y}} \pi_{\rm{data}}(\mathbf{y}| \mathbf{x})=1$. Taking the logarithm of both sides of  and using  some algebra obtains:
\begingroup\makeatletter\def\f@size{10}\check@mathfonts\def\maketag@@@#1{\hbox{\m@th\normalfont\normalfont#1}}
\begin{align}
\log \frac{\pi^{*}(\mathbf{y} \mid \mathbf{x})}{\pi_{\rm{ref}}(\mathbf{y} \mid \mathbf{x})}=f^{*}(\mathbf{x},\mathbf{y}), \label{Eq:reparameterize}
\end{align}
\endgroup
where $f^{*}(\mathbf{x},\mathbf{y})$ is the density ratio estimated by Equation~(\ref{Eq:KLIEF}) on the preference dataset. Since the optimal density ratio is now represented in terms of the optimal policy, as opposed to the discriminator model, we can explicitly derive the following maximum likelihood objective for a parameterized policy over the preference dataset \citep{rafailov2024direct}. Analogous to the approach used for density ratio estimation and using a change of variables, we can formalize the reverse KL objective as follows:
\begingroup\makeatletter\def\f@size{10}\check@mathfonts\def\maketag@@@#1{\hbox{\m@th\normalfont\normalfont#1}}
\begin{align}
\mathbb{E}_{\pi_\mathrm{chosen}} \big[- \log \pi_{\boldsymbol{\theta}} (\mathbf{y}|\mathbf{x})\big]+ \mathbb{E}_{\pi_\mathrm{ref}} \big[
    \frac{\pi_{\boldsymbol{\theta}} (\mathbf{y}_{l}|\mathbf{x})}{\pi_{\rm{ref}} (\mathbf{y}_{l}|\mathbf{x})} \big ],
\end{align}
\endgroup
Use the set of rejected responses $\mathbf{y}_{l} \sim \pi_{\rm{ref}}(\mathbf{y}\mid \mathbf{x})$ to approximate the expectations under $\pi_{\rm{ref}}(\mathbf{y}\mid \mathbf{x})$ results in our \method objective as follows:
\begin{align}
\mathcal{L}_{\rm{InfoPO}}=-\log \pi_{\theta} (\mathbf{y}_{w}|\mathbf{x})+ {\pi_{\theta} (\mathbf{y}_{l}|\mathbf{x})}/{\pi_{\rm{ref}} (\mathbf{y}_{l}|\mathbf{x})}, \nonumber
\end{align}
which completes the proof.
\end{proof}

\section{The Details of Experiments}
\subsection{Dataset Descriptions} \label{app:dataset}
\textbf{Anthropic-HH}~\citep{bai2022training}: The Anthropic Helpful and Harmless dataset\footnote{\url{https://huggingface.co/datasets/Anthropic/hh-rlhf}} contains 170,000 dialogues between humans and large language model assistants, used for single-turn dialogue evaluation tasks. Each dialogue includes a human prompt along with two model-generated responses, rated based on helpfulness and harmlessness. Consistent with DPO~\citep{rafailov2024direct}, we utilized the chosen responses during the SFT stage.

\noindent \textbf{Reddit TL;DR Summarization}~\citep{volske2017tl}: This dataset\footnote{\url{https://huggingface.co/datasets/openai/summarize\_from\_feedback}} includes forum posts from Reddit, specifically collected for summarization purposes, along with corresponding preference labels. In line with prior research~\citep{stiennon2020learning}, we employ a refined version of this dataset to train our SFT policy, leveraging its preference labels during the alignment process.

\noindent  \textbf{UltraFeedback Binarized}~\citep{cui2023ultrafeedback,tunstall2023zephyr}: This dataset\footnote{\url{https://huggingface.co/datasets/HuggingFaceH4/ultrafeedback_binarized}} comprises 64,000 prompts, each associated with four different completions produced by a mix of open-source and proprietary models. GPT-4 evaluates these completions, assigning scores based on factors such as helpfulness and honesty. Binary preference pairs are created by selecting the completion with the highest average score as the "accepted" response, while one of the other three is chosen randomly to serve as the "rejected" response.

\subsection{The Details of Experimental Setup}
\label{app:setup}
For the general hyperparameter settings, we follow the configurations established in SimPO~\cite{meng2024simpo}. Specifically, for both the SFT and preference optimization phases, we employed a batch size of 128. A cosine learning rate schedule with 10\% warmup steps was applied over a single epoch, using the Adam optimizer~\cite{kingma2014adam}. These hyperparameters were kept consistent throughout all experiments to ensure comparability. Regarding method-specific hyperparameters, we adhered to the search strategy specified in SimPO~\cite{meng2024simpo}. Each baseline model had its own set of hyperparameters, with a learning rate search range of [3e-7, 5e-7, 6e-7, 1e-6]. To counteract length bias in our methods, we normalized the response likelihood, computed as the average log probability of all tokens in the response based on the policy model, similar to the approach used in SimPO.  For SimPO and our \method, the $\beta$ in SimPO  was selected through a search within the range of [0.5, 1.0, 2.0]. For other methods, the $\beta$ search range followed a similar approach to SimPO, with values tested from [0.001, 0.01, 0.1]. All experiments were conducted on eight NVIDIA V100 32GB GPUs with a batch size of 128 based on  the alignment-handbook repository.\footnote{\url{https://github.com/huggingface/alignment-handbook}}